\documentclass[11pt,a4paper]{article}
\usepackage[hyperref]{emnlp2018}
\usepackage{times}
\usepackage{latexsym}
\usepackage{amsmath}
\usepackage{url}  
\usepackage{graphicx}  
\usepackage{subcaption,mwe}
\usepackage{comment}
\usepackage{color}
\usepackage{multirow}
\usepackage{enumitem}
\usepackage{xcolor} 

\usepackage{etoolbox}
\makeatletter
\patchcmd\@combinedblfloats{\box\@outputbox}{\unvbox\@outputbox}{}{%
   \errmessage{\noexpand\@combinedblfloats could not be patched}%
}%
 \makeatother

\aclfinalcopy 

\newcommand{\eat}[1]{}

\newcommand*{\affaddr}[1]{#1} 
\newcommand*{\affmark}[1][*]{\textsuperscript{#1}}
\newcommand*{\email}[1]{\texttt{#1}}

\title{Hierarchical Quantized Representations for Script Generation}

\author{%
Noah Weber \affmark[1], Leena Shekhar \affmark[1], Niranjan Balasubramanian\affmark[1], Nathanael Chambers \affmark[2] \\
\affaddr{\affmark[1]Stony Brook University, NY, USA}\\
\email{\{nwweber, lshekhar, niranjan\}@cs.stonybrook.edu}\\
\affaddr{\affmark[2]United States Naval Academy, MD, USA}\\
\email{nchamber@usna.edu} %
}

\date{}

\begin{document}
\maketitle

\begin{abstract}
Scripts define knowledge about how everyday scenarios (such as going to a restaurant) are expected to unfold. One of the challenges to learning scripts is the hierarchical nature of the knowledge. For example, a suspect arrested might plead innocent or guilty, and a very different track of events is then expected to happen. To capture this type of information, we propose an autoencoder model with a latent space defined by a hierarchy of categorical variables. We utilize a recently proposed vector quantization based approach, which allows continuous embeddings to be associated with each latent variable value. This permits the decoder to softly decide what portions of the latent hierarchy to condition on by attending over the value embeddings for a given setting. Our model effectively encodes and generates scripts, outperforming a recent language modeling-based method on several standard tasks, and allowing the autoencoder model to achieve substantially lower perplexity scores compared to the previous language modeling-based method.
\end{abstract}

\section{Introduction}
Scripts were originally proposed by \citet{Schank} as ``structures that describe the appropriate sequence of events in a particular context''. These event sequences define expectations for how common scenarios (such as going to a restaurant) should unfold, thus enabling better language understanding.
Although scripts represented many other factors (roles, entry conditions, outcomes) recent work in script induction \cite{Rudinger2015, Pichotta2016, Peng2016}
has focused on language modeling (LM) approaches where the ``appropriate sequence of events'' is the textual order of events (tuples of event predicates and their arguments). 
Modeling a distribution of text sequences gives the intuitive interpretation of \textit{appropriate} event sequences being roughly equivalent to \textit{probable} textual sequences.
We continue with an LM approach, but we tackle two very important LM problems that have not yet been addressed with regards to event sequence modeling.
\begin{figure}
    \centering
    \includegraphics[scale=0.3]{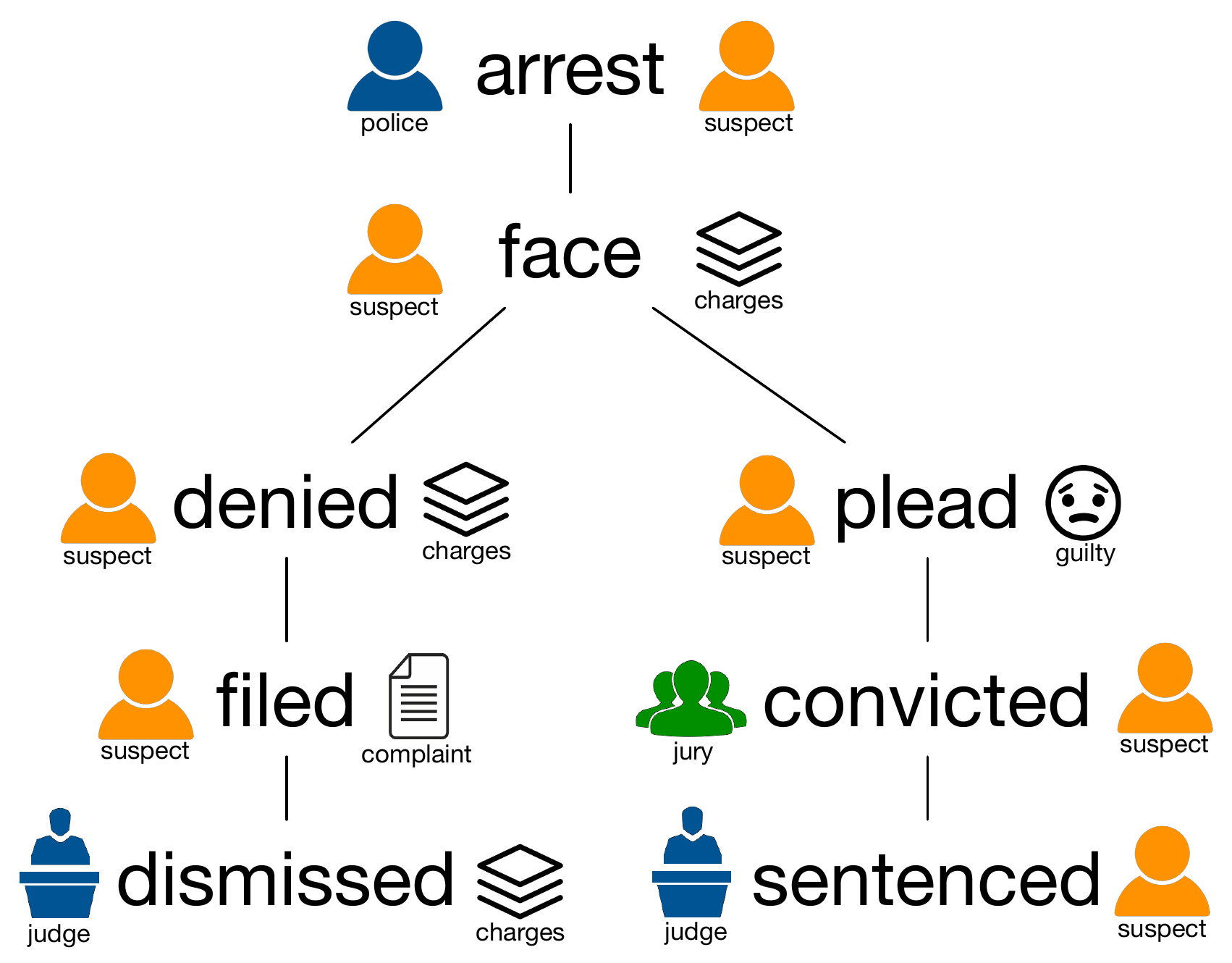}
    \caption{An automatically learned multi-track script. The left track is a dismissed case, and the right is a convicted suspect. Our model generated both tracks through a latent hierarchy.}
    \label{fig:my_label}
\end{figure}

The first problem to address is that \textit{language models tend towards local coherency}. Count based models are restricted by window size and sparse counts, while neural language models are known to rely on the local context for predictions. Since scripts are meant to describe longer coherent scenarios, this is a major issue. For example, contradictory events like \textit{(he denied charges)} and \textit{(he pleads guilty)} are given high probability in a typical language model. Our model instead captures these variations with learned latent variables.

The second problem with recent work is that \textit{the hierarchical nature of scripts is not explicitly captured}. A high level script (like a suspect getting arrested) can branch off into many possible variations. These variations are called the ``tracks" of a script. 
Figure \ref{fig:my_label} shows a script with two tracks learned by our model. LM-based approaches often fail to explicitly capture this structure, instead throwing it all into one big distribution.
This muddies the water for language understanding, making it difficult to tease apart differences like going to a \emph{fancy} restaurant or a \emph{casual} restaurant.

To remedy these problems, we propose a model that captures hierarchical structure via global latent variables. The latent variables are categorical (representing the various types of scripts and thier possible tracks and variations) and form a tree (or more generally, a DAG)\footnote{In this work we only look at linear chains of categorical variables, which is enough to encode trees (such as the one in Figure 1)}, thus capturing hierarchical structure with the top (or bottom) levels of the tree representing high (or low) level features of the script. 
The top might control for large differences like restaurant vs crime, while the bottom selects between fancy and casual dining.

The overall model, which we describe below, takes the form of an autoencoder, with an encoder network inferring values of the latents and a decoder conditioned on the latents generating scripts. We show the usefulness of these latent representations against a prior RNN language model based system \cite{Pichotta2016} on several tasks. We additionally evaluate the perplexity of the system against the RNN language model, a task that autoencoder models have typically struggled with \cite{Bowman2015}. We find that the latent tree reduces model perplexity by a significant amount, possibly indicating the usefulness of the model in a more general sense.
\section{Background}
\subsection{Variational Autoencoders}
Variational Autoencoders (VAEs, \citet{kingma2013auto}) are generative models which learn latent codes $z$ for the data $x$ by maximizing a lower bound on the data likelihood:
\[
\log(p(x)) \geq \mathrm{E}_{q(z | x)}[p(x | z)] - KL[q(z | x) || p(z)]
\]
VAEs consist of two components: an encoder which parameterizes the 
latent posterior $q(z | x)$ and a decoder which parameterizes $p(x | z)$. The objective function can be made completely differentiable via the reparameterization trick, with the full model resembling an autoencoder and the KL term acting as a regularizer. 

While VAEs have been useful in continuous domains, they have been less successful in generating discrete domains whose outputs have local syntactic regularities. Part of this is due to the ``posterior collapse" problem \cite{Bowman2015}; when VAEs are equipped with powerful autoregressive decoders, they tend to ignore the latent, collapsing the posterior $q(z | x)$ to the (usually zero-mean Gaussian) prior $p(z)$. By doing this, the model takes no penalty from the KL term, but effectively ignores its encoder. 
\subsection{Vector Quantized Variational Autoencoders}
Vector Quantized VAEs (VQ-VAEs, \citet{Oord2017}) are a recently proposed class of models which both alleviates the posterior collapse problem and allows the model to use a latent space of 
discrete values. In VQ-VAEs the latent $z$ is represented as a categorical variable that can take
on $K$ values. Each of these values $k \in \{1,...,K\}$ has associated with it a vector embedding
$e_k$. The posterior of VQ-VAEs are discrete, deterministic, and parameterized as follows:
\[
q(z=k|x) = 
\begin{cases}
1 & \text{k=}\text{argmin}_i ||f(x) - e_i||_2 \\
0 & \text{elsewise}
\end{cases}
\]
where $f(x)$ is a function defined by an encoder network.  The decoding portion of the network is similar to VAEs, where a decoder parameterizes a distribution $p(x | z=k)=g(e_k)$, where $g$ is
the decoder network, and $e_k$ is the corresponding embedding, which is fed as input to the decoder. This process can be seen as a "quantization" operation mapping the continuous encoder output to the latent embedding it falls closest to, and then feeding this latent embedding (in lieu of the encoder output) to the decoder. 

The quantization operator is not differentiable, thus during training, the gradient of
the loss with respect to the decoder input is used as an estimation to the 
gradient of the loss with respect to the encoder output. If one assumes a uniform prior over the latents (we do so here), then the KL term in the VAE objective becomes constant and may be ignored. In practice, multiple latent variables 
$z$ may by used, each with their own (or shared) embeddings space.

\section{Hierarchical Quantized Autoencoder}
\begin{figure}
    \centering
    \includegraphics[scale=0.45]{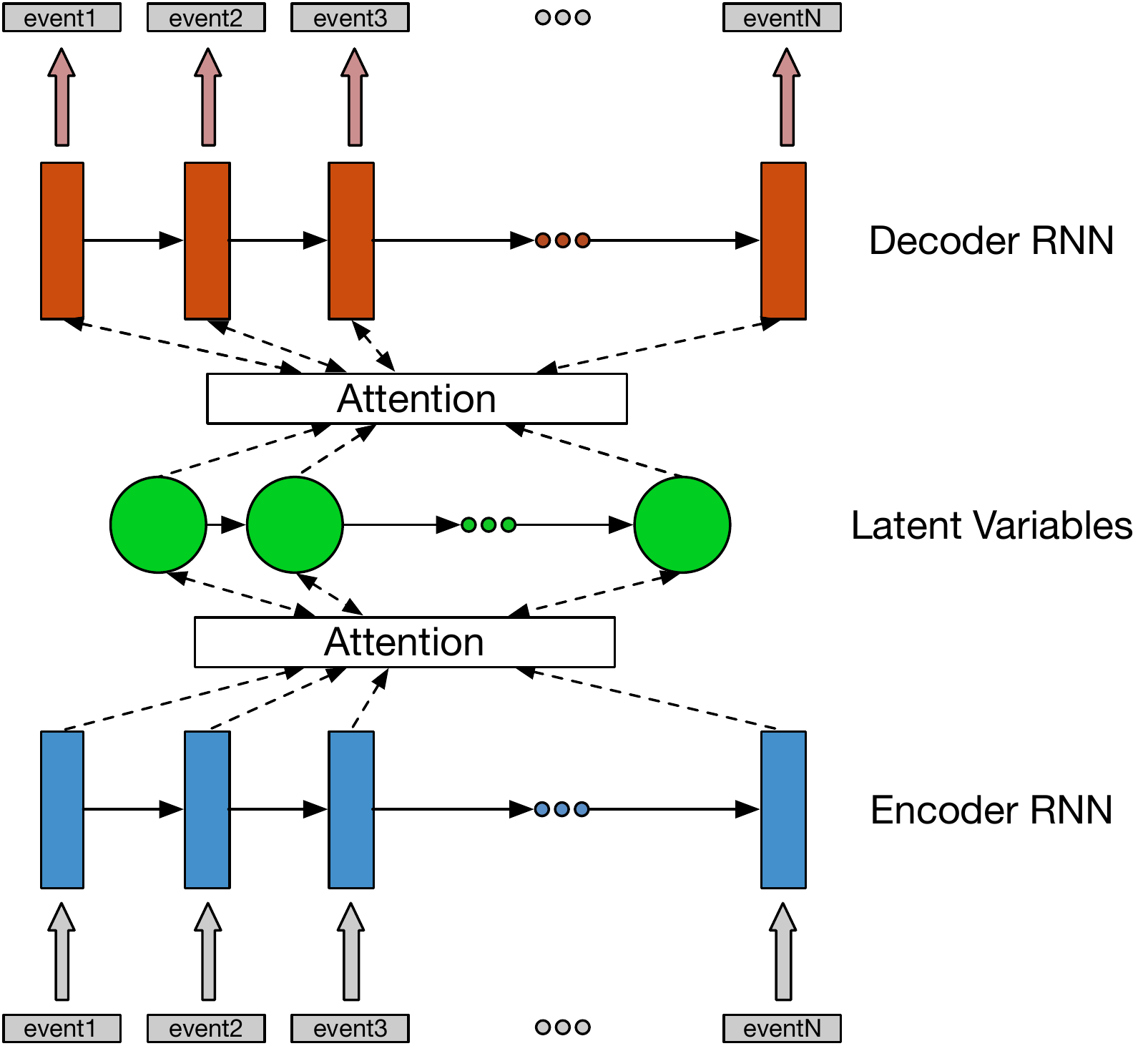}
    \caption{\label{fig:haqae}Hierarchical Quantized Autoencoder Architecture}
\end{figure}


Our goal is to build a model that can generate globally coherent multi-track scripts which allow us to account for the different ways in which a script can unfold. The main idea behind our approach is to use a hierarchical latent space which conditions the generation of the scripts. The VQ-VAE models (described earlier) provide a way to model discrete variables in an autoencoding framework while avoiding the posterior collapse problem. We build on this framework to propose a new HierarchicAl Quantized Autoencoder (HAQAE) model.

HAQAEs are autoencoders with $M$ latent variables, $z_0, ..., z_M$.
Each latent variable is categorical taking on $K$ different values. Like in VQ-VAEs, every categorical
value $k$ for variable $z$ has an associated embeddings $e_{zk}$. The latent variables are given 
a tree structure and the full posterior over all $M$ latents $\textbf{z}$ factorizes as:
\[
q(\textbf{z} | x) = q_0(z_0 | x) \prod^{M-1}_{i=1} q_i(z_i | pr(z_i), x)
\]
where $pr(z_i)$ denotes the parent of $z_i$ in the tree. Since the latent variables are meant to 
capture the hierarchical categorization of the script, we make the assumption that when
a higher level script category (for example, $z_0$) is observed with  the actual sequence of events ($x$), determining the immediate lower level category ($z_1$) is a deterministic operation. Thus, 
similar to VQ-VAEs, we parameterize the individual factors of the posterior, $q_i(z_i=k|pr(z_i), x)$, as:
\[
\begin{cases}
1 & \text{k=}\text{argmin}_j ||f_i(x, pr(z_i)) - e_{ij}||_2 \\
0 & \text{elsewise}
\end{cases}
\]
where $f_i(x, pr(z_i))$ is an encoding function specific to latent $z_i$ and $e_{ij}$ is the $j$th value embeddings for $z_i$. The distribution $p(x | \textbf{z})$
is similarly parameterized by an decoder function $g(\textbf{z}_e)$, where $\textbf{z}_e$ is the set of 
corresponding value embedding for each latent $z_i$. We describe the forms of the encoder and decoder in the next section.

\subsection{HAQAE Encoder and Decoder} 
During the encoding process, certain parts of the input may provide more evidence towards different parts of the hierarchy. For example, the event \textit{(he ate food)} gives evidence 
to the high level category of a \textit{restaurant} script, while the more specific event \textit{(he drank wine)} gives more evidence to the lower level category \textit{fancy restaurant}. Thus during encoding, it makes sense to allow each latent to decide which parts of 
the input to take into consideration, based on its parent latents. This is accomplished by parameterizing the encoding function for latent $z_i$ as an attention over the input $x$, with 
the parent of $z_i$ (more specifically, the embedding for the parent's current value) acting as the `query' vector. 
As is standard when using attention, the input sequence of events, $x=(x_1,...x_n)$, is first encoded into 
a sequence of hidden states $\textbf{h}_x=(h_1, ..., h_n)$ via a RNN encoder. The full encoding function for latent $z_i$ can thus be written as:
\[
f_i(x, pr(z_i)) = attn(\textbf{h}_x, pr(z_i))
\]
Though any attention formulation is possible, we use the bi-linear attention proposed in ~\citet{LuongPM15} in our implementations. For the root of the latent tree ($z_0$), which has no parents, we use the 
averaged value of the encoder vectors $\textbf{h}_x$ as the query vector for its attention.

We can define the decoder in a similar fashion. As is usually done, the distribution $p(x | \textbf{z}_e)$ can be defined in an autoregressive manner using a RNN decoder network. Like the encoding process, different parts of the hierarchy may affect the 
generation of different parts of the input. We thus also allow the decoder network $g(\textbf{z}_e)$
to be a RNN with a standard attention mechanism over the latent value embeddings, $\textbf{z}_e$.
Since the latent root $z_0$ is supposed to capture the highest level information about the script, we use its embedding value, (passed through an affine transformation and tanh activation) to initialize the hidden state of the decoder.
Both encoder and decoder can be trained end to end using the same gradient estimation used for VQ-VAE.

\subsection{Training Objective}
The training objective for HAQAE is nearly the same as the VQ-VAE objective proposed in \citet{Oord2017}. For a single training example, $x_i$ the objective can be written as:
\[
L = -\log{p(x_i | \textbf{z})} + \frac{1}{M}\sum^M_j L^R_j  +\frac{1}{M}\sum^M_j L^C_j
\]
where $L^R_j$ and $L^C_j$ are the \textit{reconstruct} and \textit{commit} loss for the $j$th latent variable. As in
\citet{Oord2017}, we let $sg(\cdot)$ stand for a \textit{stop gradient} operator, such 
that any term passed to it is treated as a constant.
The reconstruction loss is defined as:
\[
L^R_j = ||sg(f_j(x, pr(z_j))) - e_j^{*}||^2_2
\]
where $e_j^{*}$ is the argmin value embedding for $z_j$ (for the given input). The 
reconstruct loss is how the value embeddings for \textbf{z} are learned, and pushes 
the value embeddings to be closer to the output of the $f_i$.
The commit loss is defined as:
\[
L^C_j = \beta||f_j(x, pr(z_j)) - sg(e_j^{*})||^2_2
\]
which forces the encoder to push its output closer to \textit{some} embedding, preventing a situation in which the encoder maps inputs far away from all embeddings.
$\beta$ is a hyperparameter that weighs the commit loss\footnote{In our implementations we set $\beta=0.25$}.
Note that the commitment loss may be propogated all the way up through the hierarchy of latent nodes. We allow the latent embeddings to receive updates only from the reconstruct and 
commit loss (\textit{not} from the NLL loss).

\section{Training Details}

\subsection{Dataset and Preprocessing}

\paragraph{Dataset}
We use the New York Times Gigaword Corpus as our dataset. The corpus contains a total of around 1.8 million articles. We hold out 4000 articles from the corpus to construct our development (dev) set for hyperparameter tuning and 6000 articles for the test set. The input and output of the model is in the form of an event sequence. Each event is defined as
a 4-tuple, $(v,s,o,p)$, containing the verb, subject, object and preposition. Events without prepositions are 
given a null token in their preposition slot. The components of the events (the verb, subject, etc.) are all taken 
to be individual tokens, and can thus be treated more or less like normal text. For example, the events \textit{(he played harp), (he touched moon)}, would be tokenized and given to the model as: \textit{played he harp null tup touched he moon null}, where \textit{null} is the null preposition token and \textit{tup} is a special separation token between events.

We extract event tuples using Open Information Extraction system Ollie ~\cite{MausamOllie}. We then group together event tuples for 4 subsequent sentences to create a single event sequence. We also ignore tuples with common (\textit{is, are, be, ...}) and repeating predicates. Finally we have 7123097, 19425, and 28667 event sequences for training, dev, and test dataset respectively. For all the experiments we fix the minimum and maximum sequence lengths to be 8 and 50 respectively. 

\subsection{HAQAE Model Details}
The HAQAE model we use across all evaluations uses 5 discrete latent variables, structured in the form of a linear chain (thus no variable has more than one child or parent). Each variable can initially take on $K=512$ values, with all latents having an embeddings dimension of 256. 
The encoder RNN that performs the initial encoding of the event sequence is a bidirectional, single layer RNN with GRU cell~\cite{cho-emnlp14} with a hidden dimension of 512. The inputs to this encoder are word embeddings derived from the one-hot encodings 
of the tokens in the event sequence. The embeddings size is 300. We find initializing the embeddings with pretrained GloVe \cite{pennington2014glove} vectors to be useful.
The decoder RNN is also a single layer RNN with GRU cells with a hidden dimension of 512 and 300 dimensional (initialized) word embeddings as inputs. 
For all experiments we use a vocabulary size of 50k. We train the model using Adam \cite{kingma2014adam} with a learning rate of 0.0005, and gradient clipping at 5.0. We find that the training converges around 1.5 epochs on our dataset. Further details can be found in our implementation\footnote{\texttt{github.com/StonyBrookNLP/HAQAE}}

\subsection{Baselines}

We compare the performance of our proposed model against three previous baselines and a modification of our HAQAE model that removes explicit dependencies between latents.
\paragraph{RNN Language Model}

For our first baseline system we train a RNN sequence model. This model is 2 layered GRU cells with hidden size 512 and embedding size 300. We use Adam with a learning rate 0.001. To prevent the problem of exploding gradients, we clip the gradients at 10. We use uniform distribution [-0.1, 0.1] for random initialization and biases initialized to zero. We also use a dropout of 0.15 on the input and output embedding layers but none on the recurrent layers. We initialize the word embedding layer with pretrained Glove vectors as it improved the performance and makes the system directly comparable to HAQAE. We refer to this model as \textbf{RNNLM} in the following sections.

\paragraph{RNNLM + Role Embeddings}

We also reproduced the model from \citet{Pichotta2016} for comparison. This model is similar to the one above except that at each time step the model has an additional role marker input going into it. The marker guides the language model further by indicating what type of input is being currently fed to it: a subject, object, or predicate. These role embeddings are learned during training itself. Hyperparameters are exactly the same as the RNNLN except that the role embeddings have a dimension of 300. We will refer to this model as \textbf{RNNLM+Role}. We perform hyperparameter tuning of both the models using the development set. We use a vocabulary size of 50k. We trained both the baseline models for 2 epochs on the training set.

\paragraph{VAE}

We report results using a vanilla VAE model similar to the one used in 
\citet{Bowman2015}. The encoder/decoder for the VAE baseline has the same specs as the encoder/decoder for the HAQAE model, with a latent dimension of 300. We use linear KL annealing for the first 15000 steps and 0.5 as the word dropout rate.

\paragraph{Hierarchyless HAQAE (NOHIER)}
In order to test the effect of explicitly having a hierarchy in the latent variables, we additionally train another HAQAE model with no explicit hierarchical latent space. The model still has 5 discrete latent variables like our original model, however each of the variables are 
independent of each other (given the input). All five variables are have an attention over the input and take the average of encoder vectors $\textbf{h}_x$ as the query vector (as done with the latent root $z_0$ in the original model). We additionally 
designate one of the variables to be used to initialize the hidden state of the decoder. We found the same training hyperparmaters used in the training of the original model to work well here.

\section{Evaluation}

\subsection{Language Modeling: Perplexity}
As our proposed models are essentially language models, it is natural to evaluate their perplexity scores, which can be viewed as an indirect measure of how well the models can identify scripts. We compute per-word perplexity and per-word negative log likelihood on the validation and test sets. We compute these values without considering the end-of-sentence (EOS) token. Table~\ref{tab:perplex} gives these results. A good language model should assign low perplexity (high probability) to the validation and test sets. We observe that HAQAE achieves the minimum negative log likelihood and perplexity scores on both the validation and test sets as compared to the previous RNN-based models. The result is particularly interesting as autoencoders usually perform worse or comparable to other RNN language models in terms of perplexity (negative log likelihood) as is in the case of the vanilla VAE here; similar observations have also been made in \citet{Bowman2015}. 

\begin{table}[t!]
\centering
\begin{tabular}{c|c|c|c|c} 
 \multirow{2}{*}{System} & \multicolumn{2}{|c|}{validation} & \multicolumn{2}{c}{test}\\
 \cline{2-3} \cline{4-5}
 & NLL & PPL & NLL & PPL\\
 \hline
 RNNLM & 4.52 & 91.84 & 4.51 & 90.92 \\ 
 RNNLM+Role & 4.53 & 92.76 & 4.53 & 92.76 \\
 VAE & 4.56 & 95.58 & 4.55 & 94.63 \\
 NOHIER & 3.77 & 43.38 & 3.78 & 43.82 \\
 HAQAE & \textbf{3.73} & \textbf{41.68} & \textbf{3.74} & \textbf{42.10} \\
\end{tabular}
\caption{Negative log-likelihood (NLL) and perplexity (PPL) measures on the validation and test set. Lower is better for both metrics.}
\label{tab:perplex}
\end{table}

\subsection{Inverse Multiple Choice Narrative Cloze}

Narrative cloze evaluations of event based language models (LMs) start with a sequence of events as input and test whether the LMs correctly predict a single held-out event. The standard narrative cloze task has various issues \cite{chambers2017}. In our evaluations we opt instead for the multiple choice variant proposed in \citet{Granroth-Wilding16}.

One of our goals is to test if our generative model can produce globally coherent scripts by evaluating their ability to generate coherent event sequences. 
To evaluate this we create a new \emph{inverse narrative cloze} task. 
Instead of being given an event sequence and predicting a single event to go with it, we instead are given only \emph{one event} and the model must identify the rest of the event sequence. The model is identifying sequences of events, not single events. We chose this setup because sequences is what we ultimately want, but also because identifying single events resulted in very high scores (around 98\% accuracy). This task proved to be more challenging as an evaluation.

We thus score a model based on the probability it assigns to event sequences that begin with a single input event. A legitimate event sequence should have high probability compared to an event sequence that is stitched together using two random event sequences. We create legitimate event sequences of a fixed length (six) by extracting actual event sequences observed in documents. For every legitimate event sequence, we use the first event in the sequence as a seed event. Then, we construct detractor event sequences for this seed by appending a different sequence of events (number of events being five) from a randomly chosen document. We create five such detractor sequences for every legitimate sequence. We rank the six sequences based on the probabilities assigned by the model and then evaluate the accuracy of the top ranked sequence. A random model will uniformly choose one among the six sequences and thus will score $1\/6 = 16.60\%$ on the task. 
We report results averaged over 2000 sets of legitimate and detractor sequences.

\begin{table}[t!]
\centering
\begin{tabular}{c|c|c}
 System & validation & test\\
 \hline
 RNNLM & 25.30 & 26.30 \\ 
 RNNLM+Role & 24.60 & 26.35 \\
 VAE & 26.54 & 28.01 \\
 NOHIER & 31.68 & \textbf{34.00} \\
 HAQAE & \textbf{31.80} & 33.85 \\
\end{tabular}
\caption{Inverse narrative cloze accuracy(\%) on randomly selected 2k validation and test set. Models scored on whether they assign a higher probability to legitimate event sequence over detractor event sequences. Higher is better.}
\label{tab:inv_ncloze_acc}
\end{table}

Results in Table~\ref{tab:inv_ncloze_acc} show that the HAQAE is substantially better than both RNN LMs and vanilla VAE and similar to the NOHIER model, which shows the usefulness of the quantized embeddings overall as a global representation. The comparable results of the NOHIER model on this task might also indicate that explicitly modeling the hierarchical structure may not be completely necessary if ones only aim is to capture global coherence. The results on the perplexity task do indicate that overall, modeling the hierarchical structure is useful for better prediction.

\subsection{Comparing HAQAE and NOHIER}

Both HAQAE and NOHIER models achieve the best results across all tasks. The HAQAE model does better on the perplexity task, while results of the two models on the cloze task are nearly the same. One clear benefit of explicitly connecting the latent variables together appears to be in the efficiency of the learning. The HAQAE model performs comparable or better than the NOHIER model despite (in this case at least) having fewer parameters\footnote{NOHIER has more parameters in our case due to each latent taking a bidirectional encoder state as a query vector, as opposed to taking the parent latent vector as the query}.

\begin{figure}[t!]
    \centering
    \includegraphics[scale=0.35]{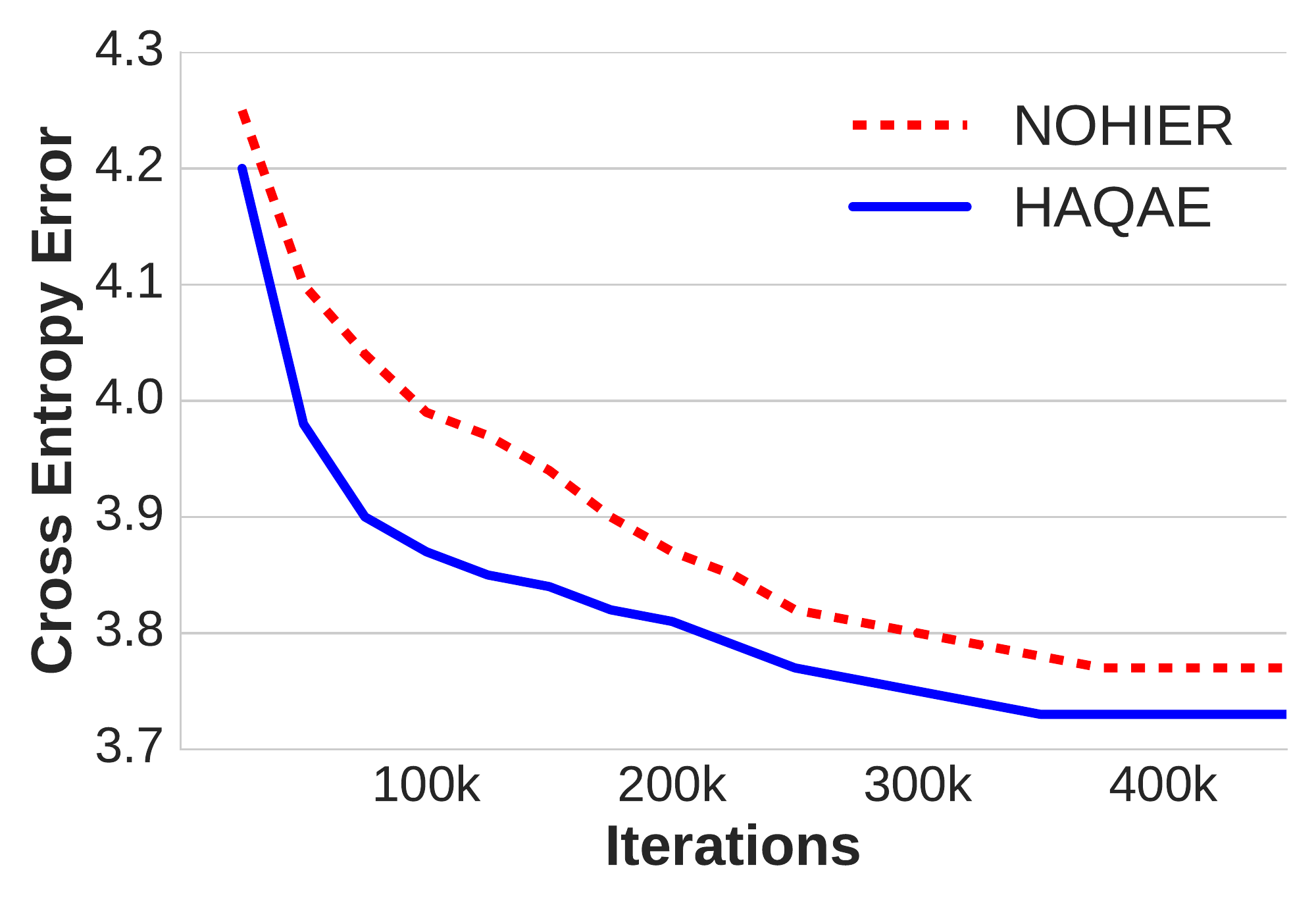}
    \caption{\label{fig:ce_error}Cross entropy error on dev set of NOHIER (\textcolor{red}{red}) and HAQAE (\textcolor{blue}{blue}) models as training progresses.}
\end{figure}

The HAQAE model also appears to learn much faster than the NOHIER model.
We show this in Figure~\ref{fig:ce_error}, which shows the per-word cross entropy error on the validation set as training progresses. We observe that the cross entropy error drops much faster in the latter model than the former one. Also, the error is always lower for the HAQAE model.

One possibility is that the NOHIER model learns similar information as the HAQAE model, but due to its lack of explicit inductive bias, takes a longer time to learn this. We leave it as future work to confirm whether this is the case through an in depth study into the properties of the learned discrete latents.

\subsection{Evaluating Event Schemas}


So far, we've evaluated how well the models recognize real textual events (perplexity) and how well the models predict events in scripts (narrative cloze). This section evaluates the script generation ability of the model, and specifically its ability to capture hierarchical information with different tracks in scripts (e.g., \emph{pleading guilty} causes different events to occur than does \emph{pleading innocent}). In many respects, this section illustrates best the power of HAQAE even though the results are partly subjective.

While we presented two automatic evaluations above, we shift to human judgment to evaluate the scripts themselves. We believe this complements the empirical gains already presented.
The scripts generated by the models were shown to human judges and scored on several metrics. 

Most previous work on script induction starts with a seed event and then grows the script based on measures of event proximity or from sampling the distribution with the seed as context.
While effective in generating a bag of events, a major problem in all previous work is that conflicting events are included (sentenced \emph{and} acquitted). While the events are related and part of the same high-level script, they should never appear together in an actual instance of a script.

In order to evaluate our model for this type of knowledge, we instead defined a seed as 2 events: the first event sets the general topic, and the second event starts a specific \textit{track} in that topic. For instance, below are two seeds that are intended to generate two tracks for the same script:

\begin{quote}
``people reported fire"\\
``fire spread in neighborhood"

``people reported fire"\\
``fire spread to forest"    
\end{quote}

For each seed (2 events in one seed) we select the first 3 events generated by a model conditioned on the seed as context. The 2 events in a seed thus initialize the latent variable values, which then inform the decoder to generate more events (we choose the first 3). The strength of our model is that the second event helps select the more specific script track, and to ignore conflicting events in other tracks.

While generating for both RNNLM+Role and HAQAE models, we additionally enforce a constraint that restricts models from generating events that have a predicate that has already been generated, as well as events whose subject and object are the same.

We evaluated the RNNLM+Role model from \citet{Pichotta2016} against our proposed HAQAE.
Each model was given 40 seeds (20 first event each with 2 contrasting second seed) and thus generated 40 scripts.
The annotators were also shown the seeds (2 events), and then asked to rate each three-event sequence for various metrics described below.

\noindent \textbf{Non-Sensical (Sense)}: Binary, is each event itself non-sensical or understandable?

\noindent \textbf{Event Relevance (Rel)}: Binary, each event was scored for being relevant or not to the script topic. This ignored whether it was consistent with the seed's branch.

\noindent \textbf{Coherency with Branch (Coh)}: 0-2, each event was scored for being coherent with the seed's specific branch (the second event). 0 means not at all, 1 means somewhat, and 2 means yes.

\noindent \textbf{Branching Uniqueness (BranchU)}: 0-2, each pair of scripts (both branches of the same topic) were scored for overlap of events. 0 means similar events generated for both, 1 means some similar events, and 2 means distinct. This score is important because some RNN decoders might ignore the second event and focus on the general topic only.

\noindent \textbf{Branching Quality (BranchQ)}: 0-2, each generated branch was scored for branch quality. 0 means the generated events are not specific to the branch, 1 means some are specific, and 2 means most/all events are specific. This is the most important score in measuring how well a model captures hierarchical structure and script tracks.

\begin{table*}
\centering
\begin{tabular}{c|c|c|c|c|c}
 System & Sense(\%) & Rel(\%) & Coh(0-2) & BranchU(0-2) & BranchQ(0-2) \\
\hline
RNNLM+Role & 93.13 & 84.73 & 1.35 & 0.65 & 0.51 \\  
HAQAE & 93.13 & 95.37 & 1.59 & 1.35 & 1.00 \\  
\end{tabular}
\caption{Human evaluation of schemas generated for the seed events. HAQAE consistently performs better than the baseline model. Higher is better for all metrics.}
\label{tab:schema_eval}
\end{table*}

\begin{table*}
\centering
\small
\begin{tabular}{|p{0.75in}|p{2.5in}|p{2.5in}|}
\hline
\textbf{RNN+Role} & \textcolor{black}{bomb found in backpack, bomb failed to detonate} & \textcolor{red}{bomb killed people, bomb detonated in blast} \\
& \textcolor{black}{bomb found in backpack, bomb detonated} & \textcolor{black}{explosion killed people, people injured in blast} \\ 
& \textcolor{black}{people reported fire, fire spread to forest} & \textcolor{red}{fire destroyed building} \\
& \textcolor{black}{people reported fire, fire spread to neighborhood} & \textcolor{black}{fire damaged building} \\
\hline
\hline
\textbf{HAQAE} &\textcolor{black}{bomb found in backpack, bomb failed to detonate} & \textcolor{blue}{they found evidence, explosive hidden in luggage} \\
& \textcolor{black}{bomb found in backpack, bomb detonated} & \textcolor{black}{blast left crater, blast killed people} \\
&\textcolor{black}{people reported fire, fire spread to forest} & \textcolor{blue}{fire burned acres} \\
& \textcolor{black}{people reported fire, fire spread to neighborhood} & \textcolor{black}{fire destroyed building} \\
\hline
\end{tabular}
\caption{Sample outputs from the baseline and our proposed system. The seeds (what is given to the system) are shown in the left column while the outputs are on the right. HAQAE is able to distinguish between the contrasting seeds. \textcolor{red}{Red} highlights the lack of branching quality in the baseline model and \textcolor{blue}{Blue} highlights the correct behavior as exhibited by HAQAE.}
\label{tab:seed-samples}
\end{table*}

\begin{table*}[h!]
\centering
\small
\begin{tabular}{|p{6in}|}
\hline
person denied charges, lawsuit filed by person, judge dismissed lawsuit  \\
person denied charges, they accused person, person resigned in january \\
\hline
clinton carried promises, clinton began in 1988, clinton made changes  \\
campaign carried promises, campaign began on september, campaign made effort \\
\hline
campaign carried promises, campaign began on september, campaign made effort \\
team carried for championship, team played in philadelphia, they won champoinship \\
\hline
\end{tabular}
\caption{Results of changing a single latent variable while keeping others fixed. Lower level latents typically change ending/beginnings or entity names (Rows 1 and 2). The top level latent changes the topic and may occasionally preserve the form (Row 3)}
\label{tab:seed-samples2}
\end{table*}

Two expert annotators evaluated the generated event sequences. In case of disagreements in scores, we also involved a third annotator to resolve these conflicts. Results for this task are shown in Table \ref{tab:schema_eval}. 

Both RNNLM and HAQAE produce sensical events, but the HAQAE model outperforms on all other metrics. It produces more relevant and coherent events for the topic at hand (relevance and coherency). But most important to the goals of this paper, it doubles the RNNLM scores on branching uniqueness and quality. This is because an RNNLM mostly generates from a bag of events after encoding the seed, but the HAQAE utilizes its latent space to produce branch-specific tracks of event sequences. Tables \ref{tab:seed-samples} show a few such examples.

\subsection{Observations about the Latent Variables}
We also look at how changing the values of various latent variables change the resulting output, in order to get a small idea as to what properties the variables capture.
We find that the root level variable $z_0$ has the largest effect on the output, and typically corresponds to the domain that the sequence of events belong to. The non root variables generally change 
the output on a smaller scale, however we find 
no correspondence between the level of the variable 
and the amount of output that is affected upon changing its value. 

One reason for the difficulty of interpreting the variables is that the model conditions on them through attention, thus changing the value of one does not necessarily need to have any effect. 

We do find that changing the lower level latents generally leads to the ending/beginning of the sequence changing or the entities of the sequence changing (but still remaining in the same topical domain). We additionally find that changing the top 
level latent may often preserve the overall form of the event sequence, and only transform the topic. We provide examples of these output by our system in Table~\ref{tab:seed-samples2}.

\section{Related Work}
Scripts were originally proposed by \citet{schank1975scripts} and further expanded upon in \citet{Schank}. The notion of hierarchies in scripts has been studied in the works of \citet{abbott1985} and \citet{bower1979scripts}. \citet{mooney} present an early non probabilistic 
system for extracting scripts from text. A highly related work by \citet{miikkulainen:recognition} provides an early example of a system explicitly designed to take advantage of the hierarchical nature of scripts, creating a model of scripts based on self organizing maps \cite{kohonen1982self}. Interestingly, self organizing maps also utilize vector quantization during learning (albeit in a different way than done here).

Recent work starting from \citet{Chambers2008} has focused on learning scripts as prototypical sequences of events using event co-occurrence. Further
work has framed this task as a language modeling problem \cite{Pichotta2016, Rudinger2015, Peng2016}. Other work has looked at learning more structured 
forms of script knowledge called \textit{schemas} \cite{Chambers2013, balasubramanian2013generating, nguyen2015generative} which focuses on additionally inducing script specific roles to be filled by entities. 
In this work we treat event components as separate tokens, though work has also looked into methods for composing this components into a single distributed event representation \cite{modi-titov:2014:W14-16, modi16, WeberTensor}. We leave this as possible future work.

The hierarchical structure of our proposed model is similar to structure of the 
latent space in other VAE variants \cite{Soenderby2016, Zhao2017}, with the discrete variables and attentions in our model being the major differences. \citet{Hu2017} present a VAE based model for controllable text generation, with different latents controlling different aspects of the generated text, but requiring labels for semi-supervision. Other methods using discrete 
variables for VAEs have also been proposed \cite{Rolfe2016}, as have variations in the VQ-VAE learning process \cite{sonderbycontinuous}

\section{Conclusion}

We proposed a new model, HAQAE, for script learning and generation that is one of the first to model the hierarchy that is inherent in this type of real-world knowledge. Previous work has focused on modeling event sequences with language models, while ignoring the problem of contradictory events and different tracks being jumbled together. The hierarchical latent space of HAQAE instead attends to the choice points in event sequences, and is able to provide some discrimination between tracks of events.

While HAQAE is motivated by the specific need for hierarchies in scripts, it can also be seen as a general event language model. As a language model HAQAE has a substantially lower perplexity on our test set than previous RNN models despite HAQAE's decoder having fewer parameters. 

We also presented a new inverse narrative cloze task that is a multiple-choice selection of event sequences. It proved to be a very difficult task with systems producing accuracies in the mid 20\% range. HAQAE and NOHEIR were the only systems to break 30 with a top accuracy of 34.0\%. This further illustrates that using a latent space to capture script differences helps identify relevant sequences.

To our knowledge, all previous work on script induction has focused on learning single event sequences or bags of events. We view our proposed model as a new step toward learning different details about scripts, such as tracks and hierarchies. Though the proposed model works well empirically, understanding exactly what is learned in the latent variables is non trivial, and is a possible direction for future work.

\subsection{Acknowledgements}
This work is supported in part by the National Science Foundation under Grant IIS-1617969. 

\bibliographystyle{acl_natbib_nourl}
\bibliography{main}
\end{document}